\documentclass [twocolumn,a4paper]{article}
\usepackage{times}
\usepackage{helvet}
\usepackage{courier}
\usepackage{amsmath}
\usepackage{graphicx}
\usepackage{booktabs }
\usepackage{subfigure}
\usepackage{amssymb}

\begin{document}

\title{T-CONV: A Convolutional Neural Network For Multi-scale Taxi Trajectory Prediction }
\author{Jianming Lv\\
South China University of Technology,Guangzhou,China\\
Email: jmlv@scut.edu.cn\\
\and
Qing Li\\
City University of Hongkong\\
Email:qing.li@cityu.edu.hk
\and
Xintong Wang\\
South China University of Technology,Guangzhou, China\\
Email:csalexwang@mail.scut.edu.cn
}
\maketitle
\begin{abstract}
Precise destination prediction of taxi trajectories can benefit many intelligent location based services such as accurate ad for passengers. Traditional prediction approaches, which treat trajectories as one-dimensional sequences and process them in single scale, fail to capture the diverse two-dimensional patterns of trajectories in different spatial scales. In this paper, we propose T-CONV which models trajectories as two-dimensional images, and adopts multi-layer convolutional neural networks to combine multi-scale trajectory patterns to achieve precise prediction. Furthermore, we conduct gradient analysis to visualize the multi-scale spatial patterns captured by T-CONV and extract the areas with distinct influence on the ultimate prediction. Finally, we integrate multiple local enhancement convolutional fields to explore these important areas deeply for better prediction. Comprehensive experiments based on real trajectory data show that T-CONV can achieve higher accuracy than the state-of-the-art methods.
\end{abstract}
 
\section{Introduction}

Taxi has become one of the major transportation tools in big cities nowadays. For efficient schedule and security monitoring of taxis running in a city, mobile GPS devices are widely installed in most of taxis to record and report their trajectories. Analysis of the destinations of taxi trajectories can benefit a lot of location based services for passengers, such as recommending sightseeing places, accurate ad based on destinations, etc.

Fig.~1 shows a typical use case of the destination prediction, in which a mobile TV installed in the running taxi reports its location to an ad company at fixed time intervals. Based on the trajectory of the taxi and the knowledge learned from the historical trajectories stored in the database, the company predicts that the destination of this trip may be close to a shopping mall with high probability, and thus it pushes some  discount information of the products currently on sale at the mall to the mobile TV. How to precisely predict the destination of a running taxi based on historical trajectories is the key problem of this kind of location based recommendation tasks.

\begin{figure}
\centering
{
\begin{minipage}[b]{0.45\textwidth}
\includegraphics[width=1\textwidth]{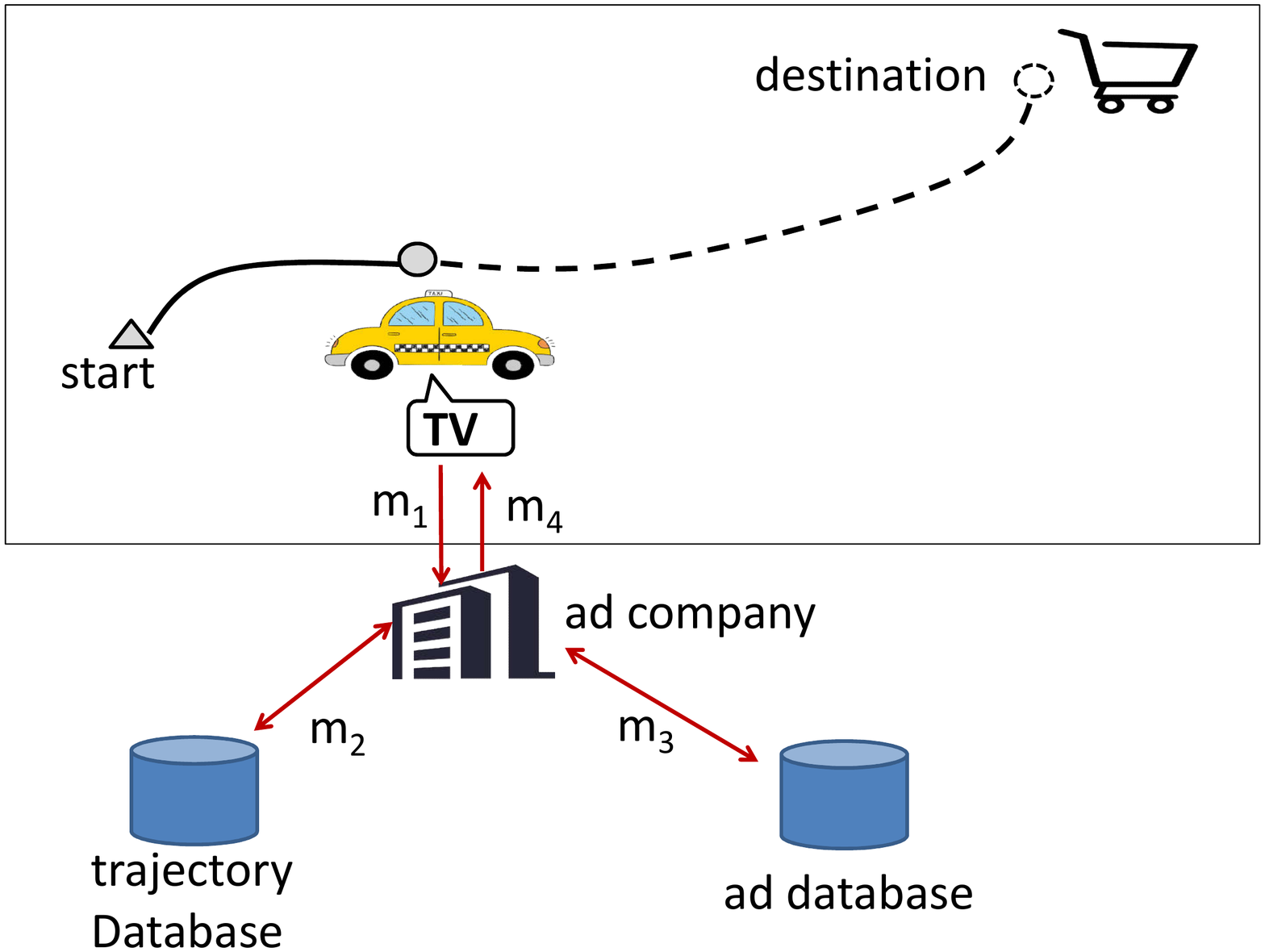}
\end{minipage}
}
\caption{A use case of the destination prediction problem: pushing ad to a running taxi based on its predicted destination. $m_1$: the running taxi reports its trajectory to the ad company. $m_2$: the company predicts that the destination of the taxi is close to a shopping mall based on the historical trajectory database. $m_3$: the company queries the ad database for the ad content related to the destination. $m_4$: the company pushes the ad to the TV terminal installed in the taxi.}
\label{sparsity}
\end{figure}

The most recent event related to this prediction problem is the ECML-PKDD competition \cite{kaggle2015}, which was set up to predict the destinations of taxi trajectories in Porto. The winner of the championship \cite{de2015artificial} adopted MLP (multi-layer perception) and RNN (recurrent neural network) algorithms. In addition, some recently proposed models \cite{xue2013destination,xue2015solving,citation-0} predict destination  by using Markov models to match query trajectories with historical records, or cluster the moving objects according to their trajectories and makes prediction according to the mobility patterns of similar objects\cite{chen2015predicting}.

Most of existing researches model trajectories as  sequences of spatio-temporal points or road segments, and process them in single spatial scale. This kind of one-dimensional data structures is hard to explicitly reveal the two-dimensional spatial features of trajectories, e.g. crossings, corners, and windings, which are highly related to the structure of road networks and  very useful to understanding of the state and destination of taxi trajectories.  Furthermore, we observe that trajectories usually have distinct multi-scale properties, which prefer a multi-scale algorithm to improve the prediction performance. In particular, trajectories in different spatial scale can exhibit different patterns. In a micro scale, a trajectory can be observed with precise positions on the road network. However, in this scale the side-effect of noise has great impact and it is hard to discover the global trend of a large spatial scope. On the contrary, in a macro scale, the global trend of a trajectory can be easily revealed, whist its resolution is relatively low and many details are lost to support accurate prediction. In fact, combinations of the patterns in different scales may help capture the motion features more precisely and completely for better prediction.

Based on the above observation, we propose a novel prediction algorithm, T-CONV, which models trajectories from a different view by considering them as two-dimensional images, and applies multi-layer convolutional neural networks (CNN)  to extract  multi-scale two-dimensional trajectory features for accurate prediction. Comprehensive experiments based on  real data \cite{kaggle2015} show that our model can achieve better precision than state-of-the-art algorithms.

The main contributions of this paper are summarized as follows:

\begin{itemize}
\item  For accurate prediction, T-CONV models trajectories as two-dimensional images and adopts multi-layer convolutional neural networks to combine multi-scale two-dimensional trajectory features. Compared with traditional one-dimensional trajectory models, T-CONV is more explicit and efficient to capture two-dimensional local patterns in different spatial scales.
\item To visualize the multi-scale patterns captured by T-CONV, the gradient distribution of features is analyzed, and the results show that T-CONV can \textcircled{\scriptsize{1}} learn small-scale trajectory patterns effectively in its lower convolutional layers, and \textcircled{\scriptsize{2}} combine them into large-scale patterns in higher layers. The visualization also reveals that the portions close to the start and the end locations of an input trajectory have distinct impact on the prediction of its ultimate destination.

\item Beyond convolving the whole input trajectory, we further integrate multiple local enhancement convolutional fields in T-CONV to further explore the specific important areas in a trajectory image, which not only  can reduce the sparsity problem in trajectory datasets, but also may greatly improve the accuracy of prediction.

\item We conduct comprehensive experiments based on real trajectory data, the results of which show that T-CONV can achieve better accuracy than state-of-the-art methods.

\end{itemize}

The following sections are organized as follows. Section II describes some related work. Section III offers problem analysis. Section IV presents T-CONV. Section V shows experimental results to validate T-CONV. We conclude this paper in Section VI.

\section{Related Work}
In this section, we introduce and review some research works related to trajectory prediction.

Most of the existing research works model a trajectory as a one-dimensional spatio-temporal sequence and predict the destination by matching the query trajectory with historical trajectories. Specifically, \cite{krumm2006predestination,wei2012constructing,ziebart2008navigate} represent a map as a two-dimensional grid in a unified and fixed spatial scale, and all spatial points in one cell are considered as one location identified by the id of the cell. In this way, each trajectory can be modeled as a sequence of cell ids. These algorithms apply the Bayesian inference method to measure the probability of a given destination conditioned on the observed partial trajectory according to the statistics of the historical records. However, in real deployment, the sparsity problem of trajectories often makes it hard to find the historical trajectories exactly matching the query one. Simply adopting Bayesian inference as \cite{krumm2006predestination,wei2012constructing,ziebart2008navigate}   may return zero probability for all candidate destinations most of the time. To solve this problem, \cite{xue2013destination,xue2015solving}  decompose trajectories into sub-trajectories connecting two adjacent locations, and adopt first-orders Markov model to infer the probability of  all candidate destinations. \cite{citation-0}  extends the model to consider the difference between destinations and passing-by locations by adopting an absorbing Markov chain model. \cite{kim2007path}  utilizes the road network information and divides trajectories into road segments; the research aims to predict the future path of  a moving object by matching the query trajectory with historical trajectories on a road network.  \cite{chen2010system}  expresses a personal trajectory as a sequence of cells in a grid, and organizes movement patterns in a pattern tree to support prediction.

Furthermore, clustering algorithms are commonly used techniques to overcome the sparsity problem and extract meaningful basic units in trajectories. \cite{chen2015predicting} clusters the moving objects according to their trajectories and makes prediction using the mobility patterns of similar objects. \cite{besse2016destination} clusters historical trajectories and uses the Gaussian mixture model to describe the distribution of spatial points in a trajectory cluster. Each query trajectory is assigned to multiple clusters and the mean of the destinations of these clusters is the predicted destination. \cite{alvarez2010trip} uses clustering algorithms to extract some important spatial points, named the support points, which are close to crossroads on trajectories. An HMM based algorithm is adopted to establish the relationship between support points and destinations. \cite{yang2014predicting} proposes a variant of the DBSCAN clustering algorithm to obtain stay points, and applies the variable order Markov Model to predict next locations of personal trajectories.

More recently,  neural network based algorithms are receiving more attentions, as they can more accurately predict destinations of taxi trajectories; one such algorithm \cite{de2015artificial} actually won the championship of the ECML-PKDD competition \cite{kaggle2015}. As reported, \cite{de2015artificial} considers the first k points from the start and the last k points close to the end of a query trajectory, and feeds these spatial points into neural networks to  perform prediction. A lot of neural networks are tested in the research, including normal multi-layer perception, LSTM\cite{hochreiter1997long}  based RNN(recurrent neural network), bi-directional RNN, and memory networks. Experiments based on a large validation set show that the bi-directional RNN outperforms the other methods.

Most of above studies extract the features of trajectories in only one specific spatial scale. Specifically, in the grid based algorithms \cite{krumm2006predestination,wei2012constructing,ziebart2008navigate,xue2013destination}, the scale is determined by the density of the grid, which is usually predefined and unified in the systems. The clustering  based algorithms \cite{chen2015predicting,besse2016destination,alvarez2010trip,yang2014predicting} cluster the spatial points and trajectories through their spatial distribution, and perform destination prediction based on the clusters which form a specific kind of spatial scale and  keep unchanged during the prediction. On the other hand,  in the neural network based models such as \cite{de2015artificial}, trajectories are processed in  the finest granularity, and prediction is conducted based on the spatial points, each of which is represented as a tuple of latitude and longitude.

\section{Problem Analysis}
\subsection{Problem Definition}

A taxi has two different types of states: the idle state (no passengers) and the occupied state (with passengers), which can be distinguished by the signals collected from the seat occupancy sensors in the taxi. As shown in the $m_1$ and $m_2$ steps of Fig.~1, the problem considered in this paper is to predict the destination of a running taxi in the occupied state.

Formally, given a taxi $T_i$, its no. $j$ trajectory $\zeta_{i,j}$ is usually recorded as a sequence of GPS locations collected from the GPS devices at fixed time intervals:
 \begin{eqnarray}
 \zeta_{i,j} = <\delta_{i,j,1}, \delta_{i,j,2},...,\delta_{i,j,N_{i,j}}>.
 \end{eqnarray}
Here
\begin{eqnarray}
\delta_{i,j,k} = (\Phi_{i,j,k}, \Upsilon_{i,j,k})(1\leq k \leq N_{i,j})
\end{eqnarray}
is a GPS location. $\Phi_{i,j,k}$ and $\Upsilon_{i,j,k}$ are the longitude and latitude of  $\delta_{i,j,k}$.  $N_{i,j}$ is the length of the trajectory $\zeta_{i,j}$.  Especially, the first location $\delta_{i,j,1}$ means the start of the current trip, which indicates where the passenger(s) came from. The last location $\delta_{i,j,N_{i,j}}$ is the newest location of the taxi. The ultimate destination of any trajectory $\zeta_{i,j}$ is denoted by $\Theta(\zeta_{i,j})$.

The prediction problem can be defined as follows: given any trajectory  $\zeta_{i,j}$ of a running taxi $T_i$, we need to predict the ultimate destination $\Theta(\zeta_{i,j})$ of the taxi in the current trip based on the knowledge learned from the historical trajectory set.

\begin{figure}
\centering
\subfigure[]{
\begin{minipage}[b]{0.2\textwidth}
\includegraphics[width=1\textwidth]{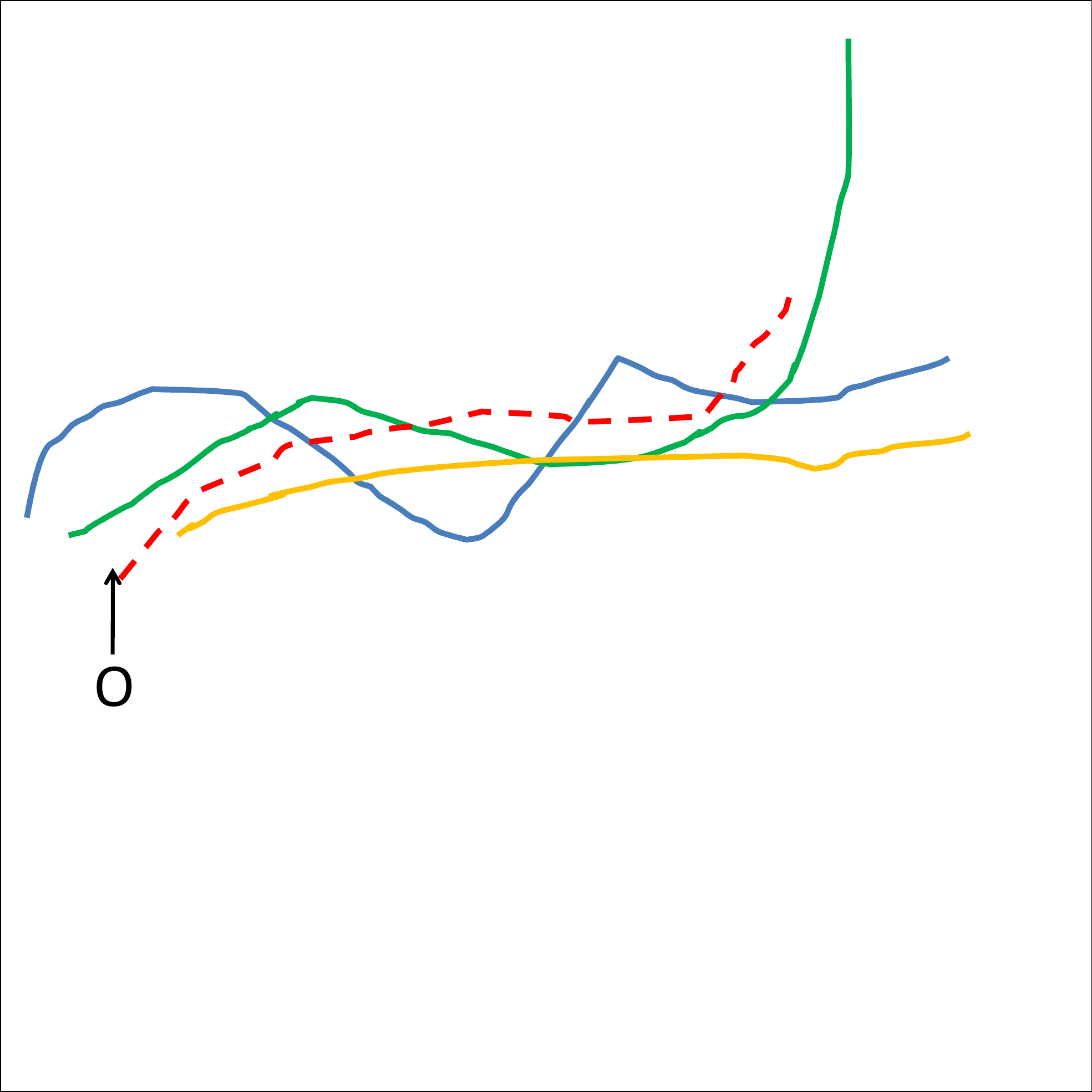}
\label{scale0}
\end{minipage}
}
\subfigure[]{
\begin{minipage}[b]{0.2\textwidth}
\includegraphics[width=1\textwidth]{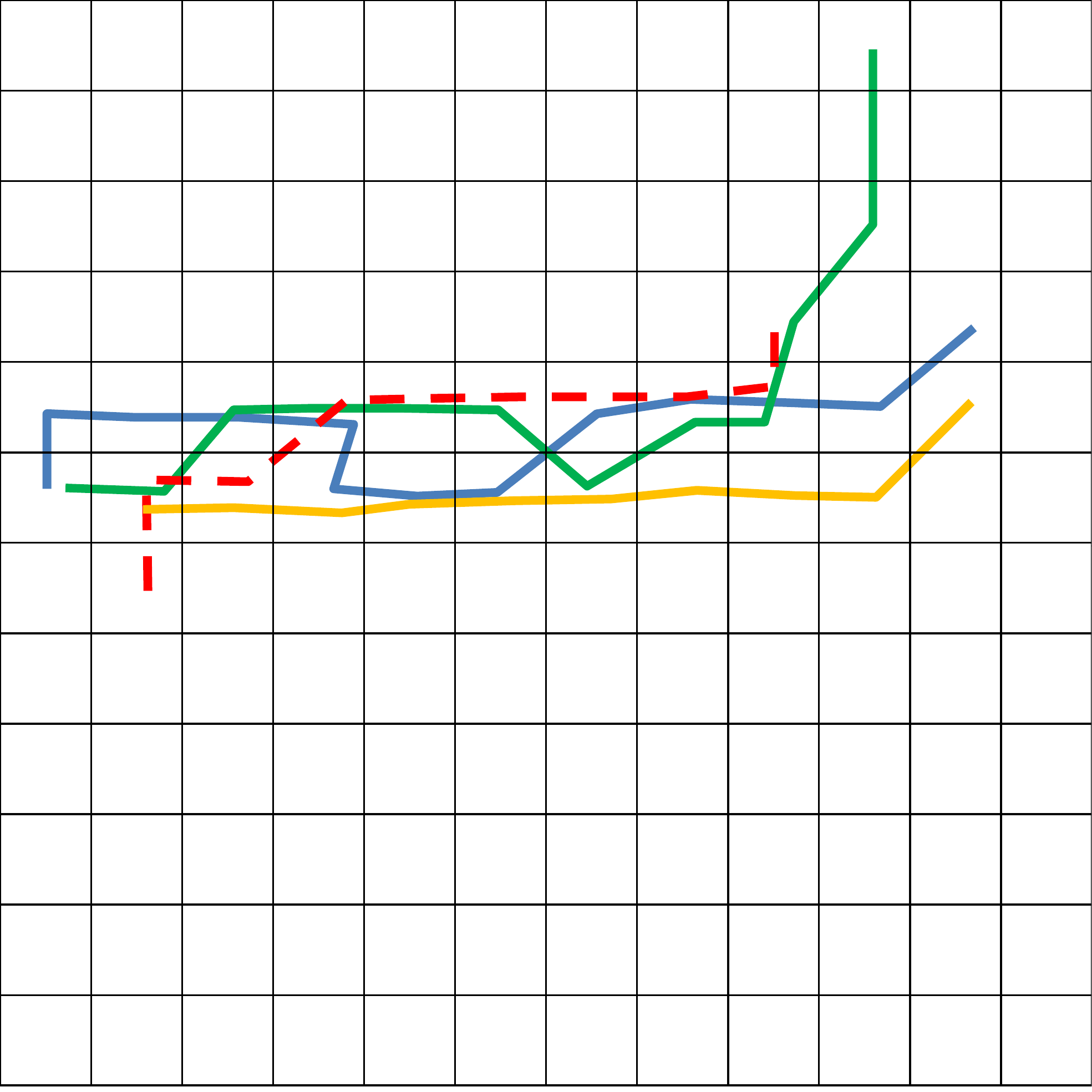}
\label{scale1}
\end{minipage}
}
\subfigure[]{
\begin{minipage}[b]{0.2\textwidth}
\includegraphics[width=1\textwidth]{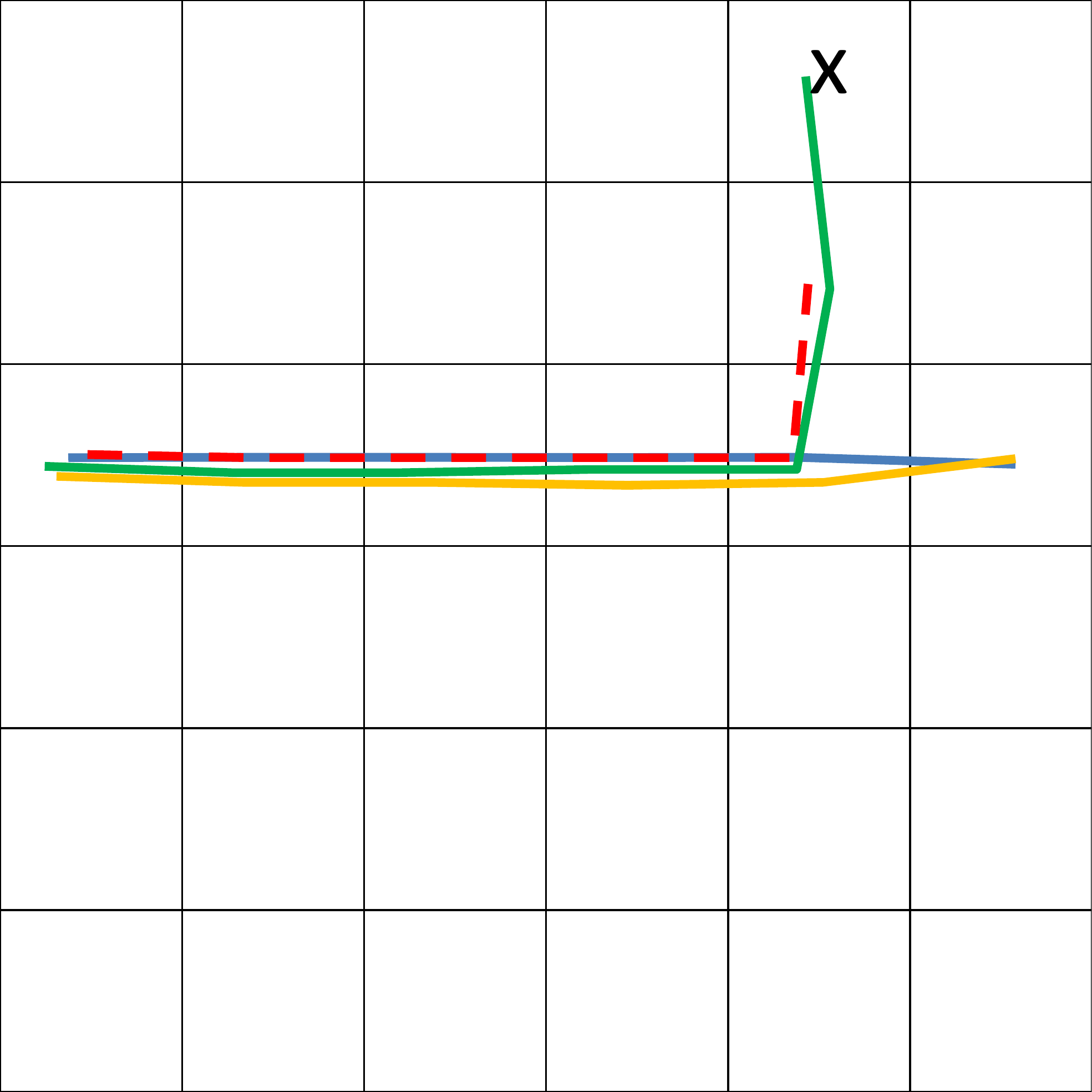}
\label{scale2}
\end{minipage}
}
\subfigure[]{
\begin{minipage}[b]{0.2\textwidth}
\includegraphics[width=1\textwidth]{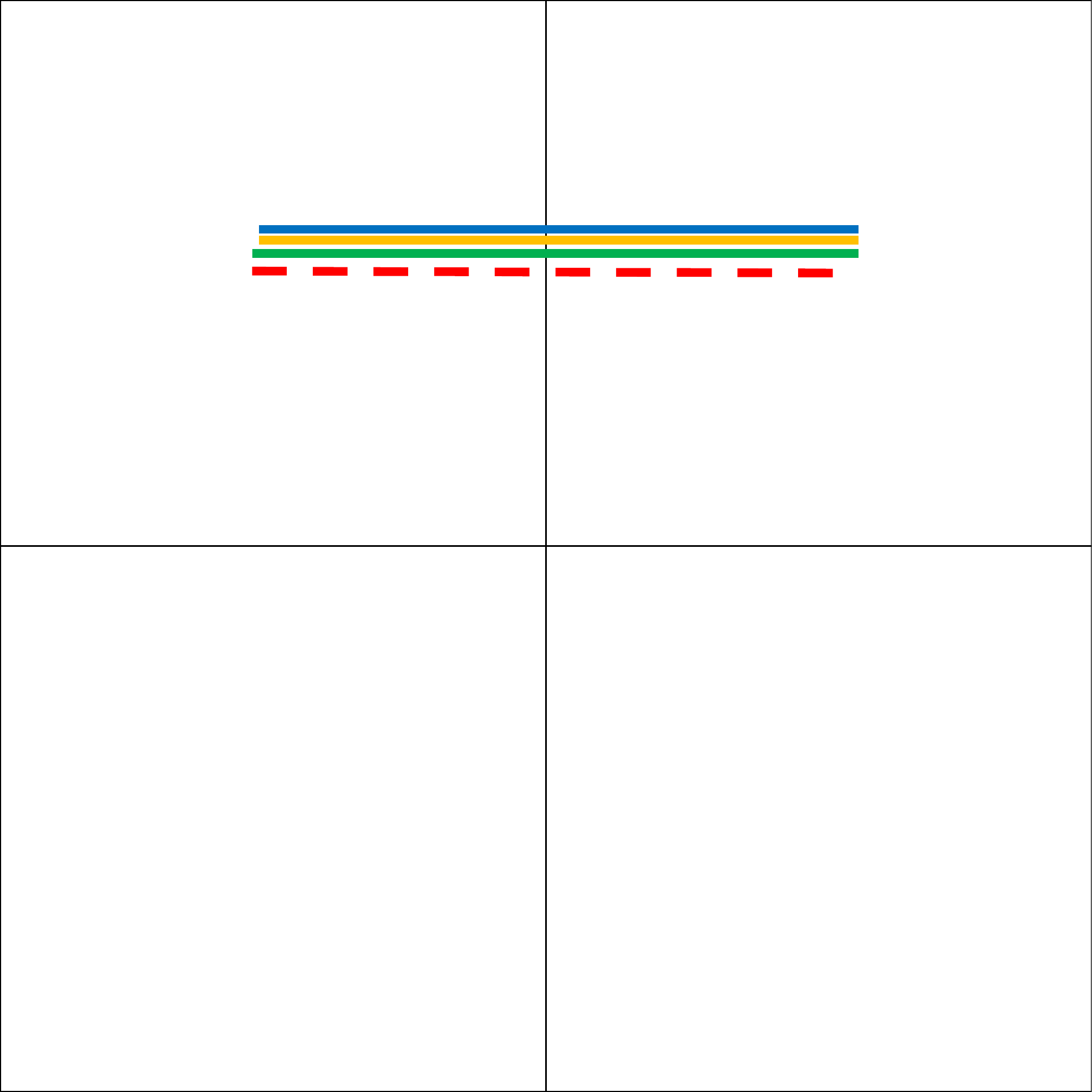}
\label{scale3}
\end{minipage}
}
\caption{Trajectories in different spatial scales. (a) Original trajectory. (b-d) Trajectories showed in the grids of different spatial scale.}
\label{multiscale}
\end{figure}

\subsection{Multi-scale Properties of Trajectories}

While trajectories are displayed on the map, a two-dimensional image such as Fig.~\ref{scale0} can be obtained. Similar to other two-dimensional spatial data such as weather and traffic data, trajectories also have significant multi-scale properties: in different spatial scales, trajectories can exhibit different two-dimensional patterns. Fig.~\ref{multiscale} shows a typical example of taxi trajectories in different scales. The dash line represents the trajectory $\zeta_{i,j}$ of a running taxi starting from the location marked with 'O'.  Other trajectories in the figures are historical trajectories in the training set. Fig.~\ref{scale0} shows the original trajectories in the highest resolution. When processing trajectories in different scales, the grid is a commonly used tool to divide the map and combine the spatial points belonging to the same cell into one indivisible unit. Fig.~\ref{scale1} shows the micro spatial scale with a dense grid, in which trajectories are recorded with fine granularity. In this scale, the view of trajectories displays most details of their motion patterns, and the overlapping degree of the trajectories is low, which causes the sparsity problem when performing trajectory matching.  As we divide the map into a sparser grid, we can illustrate the trajectories in a larger spatial scale, as shown by Fig.~\ref{scale2}. In this scale, the overlapping of the trajectories increases and the global trend of moving objects are easier to be captured. The trajectories are grouped into the two clusters, according to which the destination of the taxi looks like to be 'x' with a high probability. Fig.~\ref{scale3} shows the macro spatial scale, in which the global trend of trajectories is clearer and all trajectories overlap, but no detail about the local difference is preserved, which may nevertheless be important for accurate prediction.

In a nutshell, trajectories in small scales show micro local spatial patterns like  corners, curves and crossings, but cause sparsity problem. On the other hand, in macro scales they show the global trend clearly but it is difficult to sense important local  changes. A proper combination of the trajectory patterns in multiple scales may be better than the existing single scale based methods to capture both the micro and macro spatial patterns for more accurate prediction.

\section{Trajectory Modeling and Prediction}
\subsection{Image Based Modeling of Trajectory}

In order to extract the multi-scale two-dimensional spatial patterns explicitly, we  model trajectories as two-dimensional images instead of traditional one-dimensional sequences \cite{de2015artificial,xue2013destination,xue2015solving,citation-0,chen2015predicting}.

Firstly, we divide the map into an $M*M$ grid evenly as in Fig.2(b), where $M$ is a predefined constant representing the resolution of the map. We use  $C_{m,n}(1 \leq m,n \leq M)$ to denote the cell in the row $m$ and column $n$ of the grid. Each GPS point $\delta_{i,j,k} = <\Phi_{i,j,k},\Upsilon_{i,j,k}>$ can be mapped to a cell $C_{x,y}$ according to its latitude and longitude. The mapping relationship is denoted as:
\begin{eqnarray}
\delta_{i,j,k} \triangleright C_{x,y}.
\end{eqnarray}
In this way, given a trajectory $\zeta_{i,j}$ we can  achieve a two-dimensional $M*M$ image $I_{i,j}$, where the value of the pixel $I_{i,j}(m , n)(1 \leq m,n \leq M)$ is defined as:
\begin{flalign}
&I_{i,j}(m,n)& \\\nonumber
&\quad  =\begin{cases}
f(\delta_{i,j,k})&\text{if  $\exists \delta_{i,j,k}  (\delta_{i,j,k} \in \zeta_{i,j} \wedge \delta_{i,j,k} \triangleright C_{m,n} )$ } \\
 0&\text{ otherwise }
\end{cases}&
\end{flalign}
The image  $I_{i,j}$ is called the \emph{trajectory image} of $\zeta_{i,j}$ in this paper. Here, the function $f(\delta_{i,j,k})$ returns a non-negative value for the GPS point $\delta_{i,j,k}$ to indicate the location information.

\subsection{Prediction Based on CNN}
Inspired by the successful application of the convolutional neural networks (CNN) based algorithms \cite{lecun1998gradient,krizhevsky2012imagenet}, which enable extracting multi-scale features from pixel-level raw data of images, we propose a CNN based model, T-CONV, to capture the multi-scale spatial patterns from trajectory images and combine these patterns for destination prediction.

\begin{figure*}
\centering
{
\begin{minipage}[b]{0.58\textwidth}
\includegraphics[width=1\textwidth]{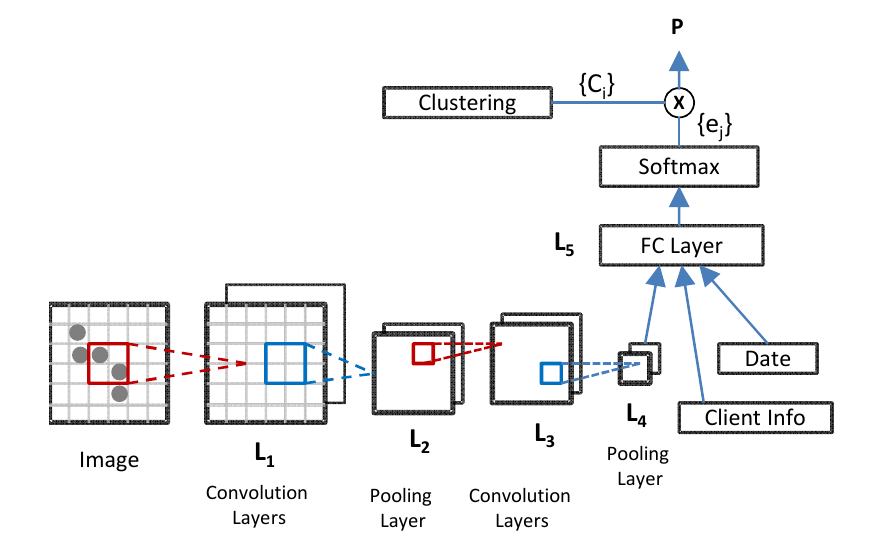}
\end{minipage}
}
\caption{The architecture of the T-CONV model based on CNN.}
\label{cnn}
\end{figure*}

The model is shown in Fig.~\ref{cnn}. Given a query trajectory $\zeta_{i,j}$, it is transformed into an $M*M$ trajectory image $I_{i,j}$ and input into the model. The value of the pixel $I_{i,j}(m , n)(1 \leq m,n \leq M)$ is defined as:
\begin{flalign}
&I_{i,j}(m,n)& \\\nonumber
&\quad =\begin{cases}
0&\text{ if $\nexists \delta_{i,j,k}  (\delta_{i,j,k} \in \zeta_{i,j} \wedge \delta_{i,j,k} \triangleright C_{m,n} )$} \\
1&\text{ if $\delta_{i,j,N_{i,j}} \triangleright  C_{m,n} $ }\\
0.5&\text{ otherwise }
\end{cases} &
\end{flalign}
In order to distinguish the end point of the trajectory from other points, $I_{i,j}(m,n)$ is set to 1 when the last point $\delta_{i,j,N_{i,j}}$ of the trajectory is in $C_{m,n}$, and it is set to 0.5 for other points in the trajectory.

The input images are fed into the CNN model, which is composed of  groups of convolution layers and max-pooling layers. Specifically, the convolution layer $L_{1}$ convolves $I_{i,j}$ to generate the multi-channel feature map $F^{(1)}_{C\times M \times M}$ (with $C$ channels), each element of which is:
\begin{small}
\begin{flalign}
&F^{(1)}_{c,m,n} & \\ \nonumber
&\quad = \sigma(\sum_{h= - \lfloor k/2 \rfloor }^{\lfloor k/2 \rfloor} \sum_{w= - \lfloor k/2 \rfloor }^{\lfloor k/2 \rfloor} \theta_{c,h,w} \cdot I_{i,j}(m+h,n+w) + b_c ).&
\end{flalign}
\end{small}

Here $\theta_{c,h,w}$ is the convolution parameter, $b_c$ is the bias parameter, $k$ indicates the size of the sliding convolution window, and $\sigma(.)$ is the activation function, typically a rectified linear activation RELU: $\sigma(x) = max(x,0)$. The main contribution of the convolution operations is to extract the local spatial dependency among neighboring spatial points.

The following max-pooling layer $L_{2}$ in Fig.~\ref{cnn} down-samples the input feature maps to generate high-level features in a larger scale. It filters out the pixels with less activation values to generate distilled feature maps $F^{(2)}_{C \times M/t\times M/t}$ , each element of which is:
\begin{eqnarray}
F^{(2)}_{c,m,n} = \max_{ 0 \leq h,w \leq t-1} F^{(1)}_{c,m*t+h,n*t+w}
\end{eqnarray}
Here $t$ indicates the pooling size.

The combination of the convolution and max-pooling operations can be applied to the feature maps multiple times to form a hierarchical CNN, such as the layers $L_1 \thicksim L_4$ in Fig.~\ref{cnn}. Higher-level features extract the patterns in larger scales, which are calculated from the combination of lower-level features in smaller scales. In this way, multi-scale patterns are embedded into the output features of the highest level of the CNN.

The output features of $L_4$ are then flattened into one-dimensional vector $S$ and fed to the fully connected neural layer $L_5$. Moreover, the meta data about date and client information are  transformed as embedded vectors \cite{de2015artificial} and also flattened into one-dimensional vector $S'$, which is input to the layer $L_5$. Each output element of  $L_5$ is calculated as follows:
\begin{eqnarray}
e_i = tanh(\sum_j W_{i,j} \times S_j + W'_{i,j} \times S'_j+ b_i)
\end{eqnarray}
Here $W_{i,j}$, $W'_{i,j}$, and $b_i$ are the parameters of $L_5$, which need to be learnt.

In the final steps, similar with  \cite{de2015artificial}, we calculate the predicted desptination $P$ as the weighted sum of the  frequently visited locations, which are the cluster centers achieved by the mean-shift clustering algorithm deployed on the destinations of all training trajectories. In particular, we have:
 \begin{eqnarray}
P= \sum_i c_i \times \frac{exp(e_i)}{\sum_j exp(e_j)}.
 \end{eqnarray}
Here $\{c_i\}$ is the set of cluster centers, and the weight on each cluster center is determined by the softmax function of the output elements $\{e_i\}$ of $L_5$.

\subsection{Visualization and Distillation of Patterns}
In order to deeply understand the semantics of the features captured by the convolutional neural network, we can measure the relationship between the output features and the input trajectory image by calculating the following gradient:
\begin{eqnarray}
g(k,c,m,n,x,y) = \frac{\partial F^{(k)}_{c,m,n}}{\partial I(x,y)}
\end{eqnarray}
Here $F^{(k)}_{c,m,n}$ indicates an output feature of the layer  $L_k (1 \leq k \leq 4)$, and $I(x,y)$ is a pixel of the trajectory image. The gradient indicates the contribution of  the pixel value $I(x,y)$ on the feature $F^{(k)}_{c,m,n}$. By visualizing the gradient distribution of all pixels in the image, we can observe what patterns captured by the feature $F^{(k)}_{c,m,n}$ from the raw input image. This  visualization method is similar with the Deconvolution Network presented in  \cite{zeiler2014visualizing}.

\begin{figure*}
\centering
{
\subfigure[]
{
\begin{minipage}[b]{0.3\textwidth}
\includegraphics[width=1\textwidth]{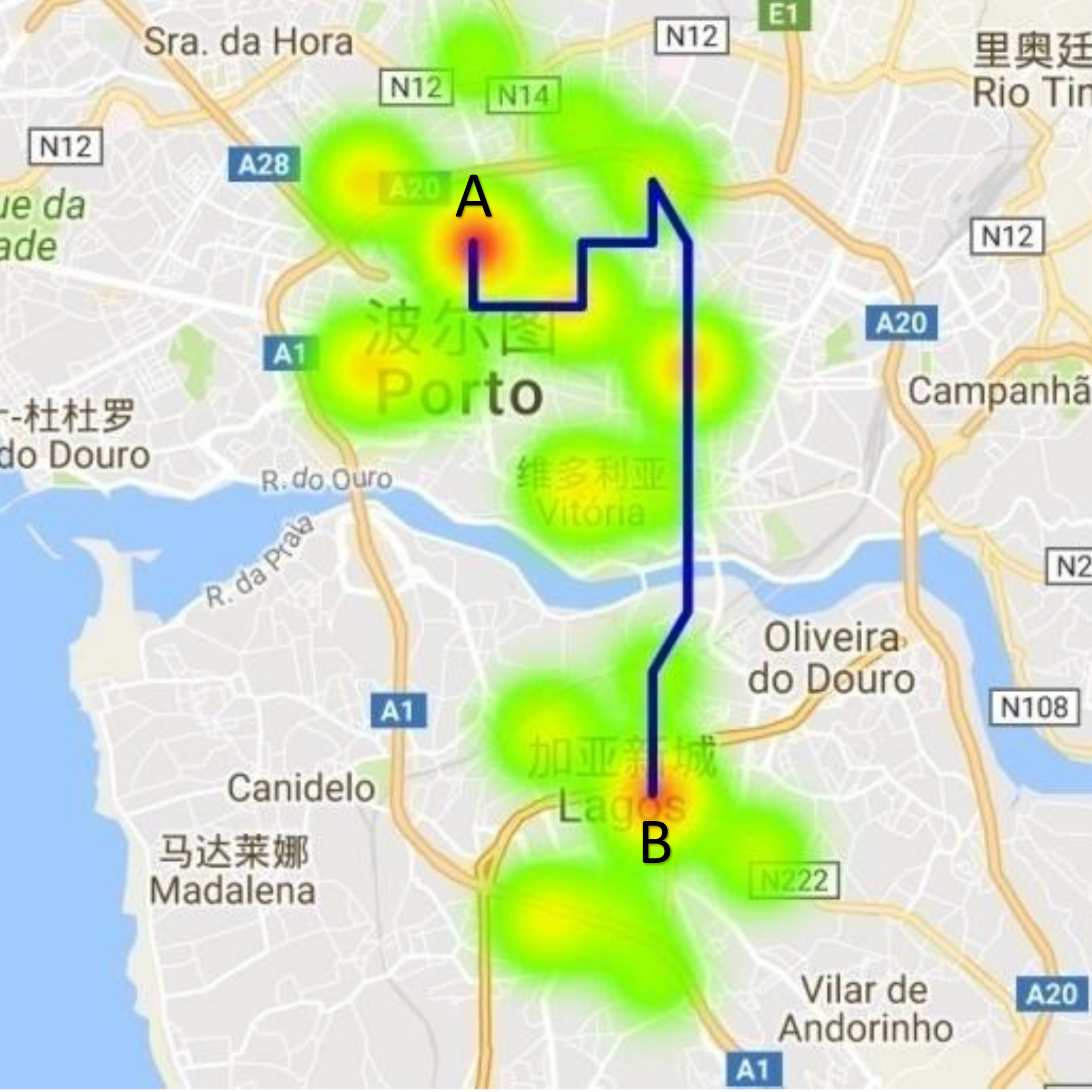}
\label{L3-pattern}
\end{minipage}
}
\subfigure[]
{
\begin{minipage}[b]{0.3\textwidth}
\includegraphics[width=1\textwidth]{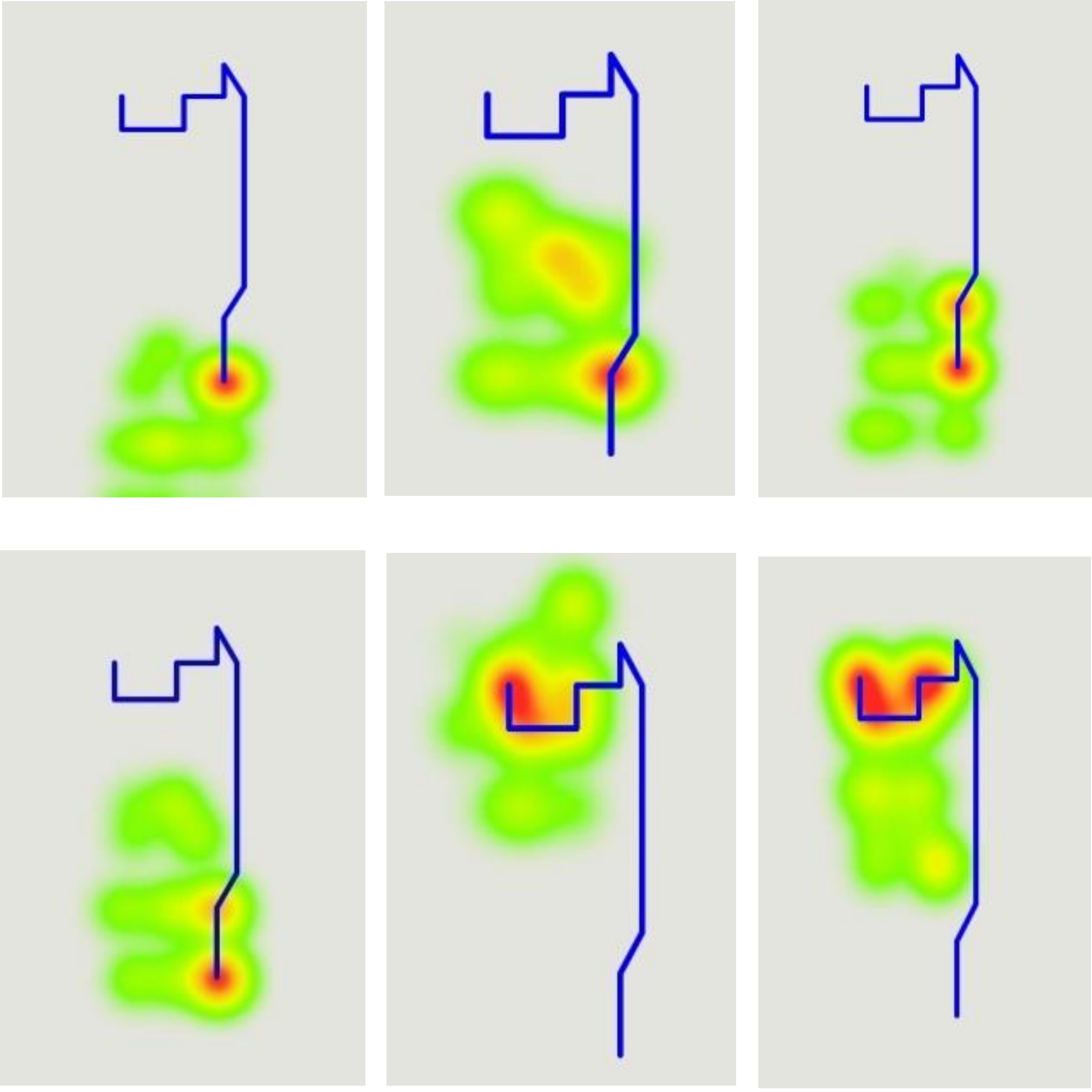}
\label{L1-pattern}
\end{minipage}
}
}
\caption{(a) The patterns captured in the layer $L_{4}$ of T-CONV. The orientation point of the trajectory is the 'A' point and the end point is 'B'. (b)The pattern captured in the layer $L_{2}$ of T-CONV.}
\label{pattern}
\end{figure*}

We validate this idea by training the model with the real trajectory set from the ECML-PKDD competition \cite{kaggle2015}, and visualize the patterns captured by the features in the different layers of T-CONV.  Fig.~\ref{L3-pattern} shows the patterns captured by the convolutional layer $L_{4}$ while processing a typical trajectory. This pattern is achieved by firstly selecting the largest output feature of $L_{4}$ which has the greatest impact  on the prediction result, then calculating the gradients over each pixel by Eq.~(10), and showing the gradient distribution on the map. In a similar way, as shown in Fig.~\ref{L1-pattern}, we select the top 6 output features of the layer $L_{2}$ and visualize the patterns captured by them.

Fig.~\ref{L1-pattern}  shows that layer $L_{2}$ can capture the small-scale patterns, which focus on the detailed changing in small local areas. Meanwhile, Fig.~\ref{L3-pattern}  shows that the higher layer $L_{4}$ can combine the lower-level small-scale patterns  into effective patterns in a larger scale. Furthermore, the red and thick areas in Fig.~\ref{pattern}  correspond to bigger gradients and indicate  important portions in the patterns. From Fig.~\ref{L3-pattern}, it is interesting to find that the trained T-CONV model can "recognize" that the local areas close to the start point and end point of the trajectory contribute much more to the destination prediction. Intuitively, the end part of a trajectory is important because it is closest to the destination, and it reveals the latest trend of the trajectory.  Meanwhile, the start portion of a trajectory is also important, as it indicates where the customers come from and represents some inner motivation of the trip.

\begin{figure}
\centering
{
\begin{minipage}[b]{0.45\textwidth}
\includegraphics[width=1\textwidth]{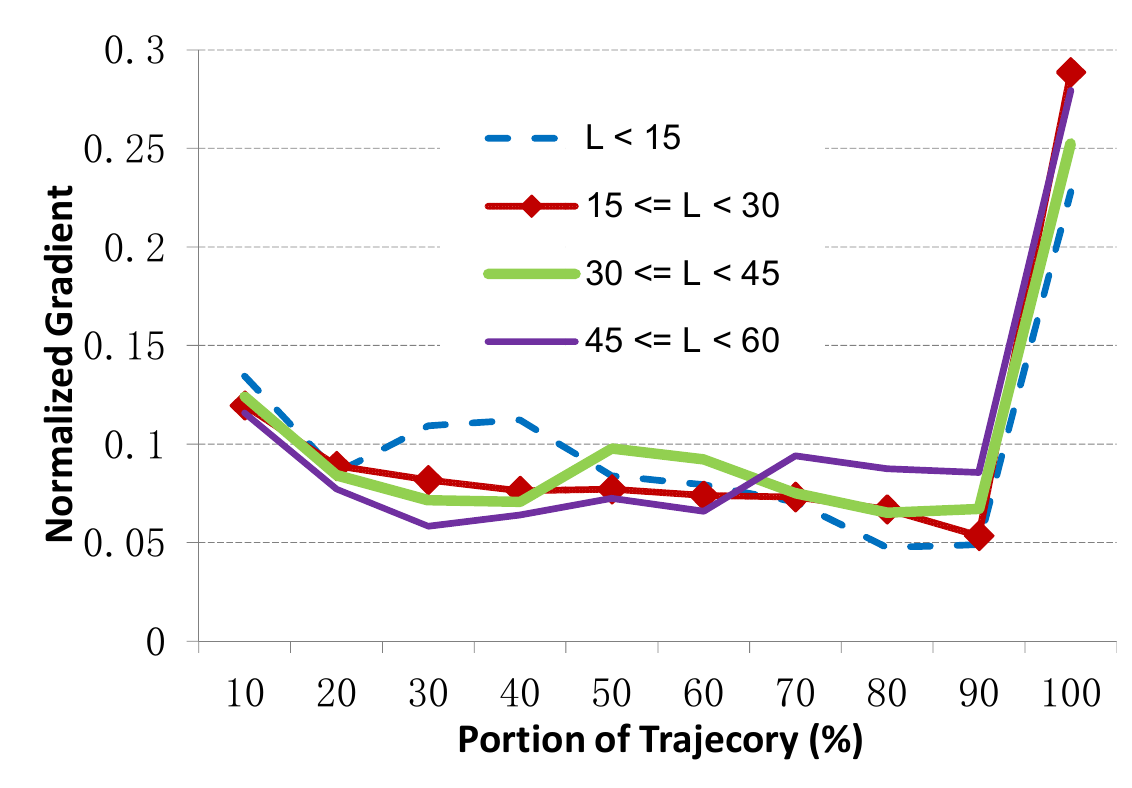}
\end{minipage}
}
\caption{The gradient distribution of different portions of a trajectory. Here $L$ denotes the time duration of the trajectories in minutes.}
\label{gradient}
\end{figure}

To further verify the uneven contribution of different portions in a trajectory, we randomly select 1,000 trajectories from the dataset \cite{kaggle2015} and calculate the gradient distribution of different portions of a trajectory as  shown in Fig.~\ref{gradient}. The distribution is achieved by calculating the gradients of  the largest feature in $L_4$ over the pixels in each trajectory, and make statistics of the gradients in different portions of a trajectory. It shows clearly that the first $10\%$ and last $10\%$ portions of a trajectory have great impact on the destination prediction. This observation motivates us to design better prediction models by  further exploring these two important local areas.

\subsection{Local Enhancement CNN for Prediction}
\begin{figure}
\centering
{
\begin{minipage}[b]{0.45\textwidth}
\includegraphics[width=1\textwidth]{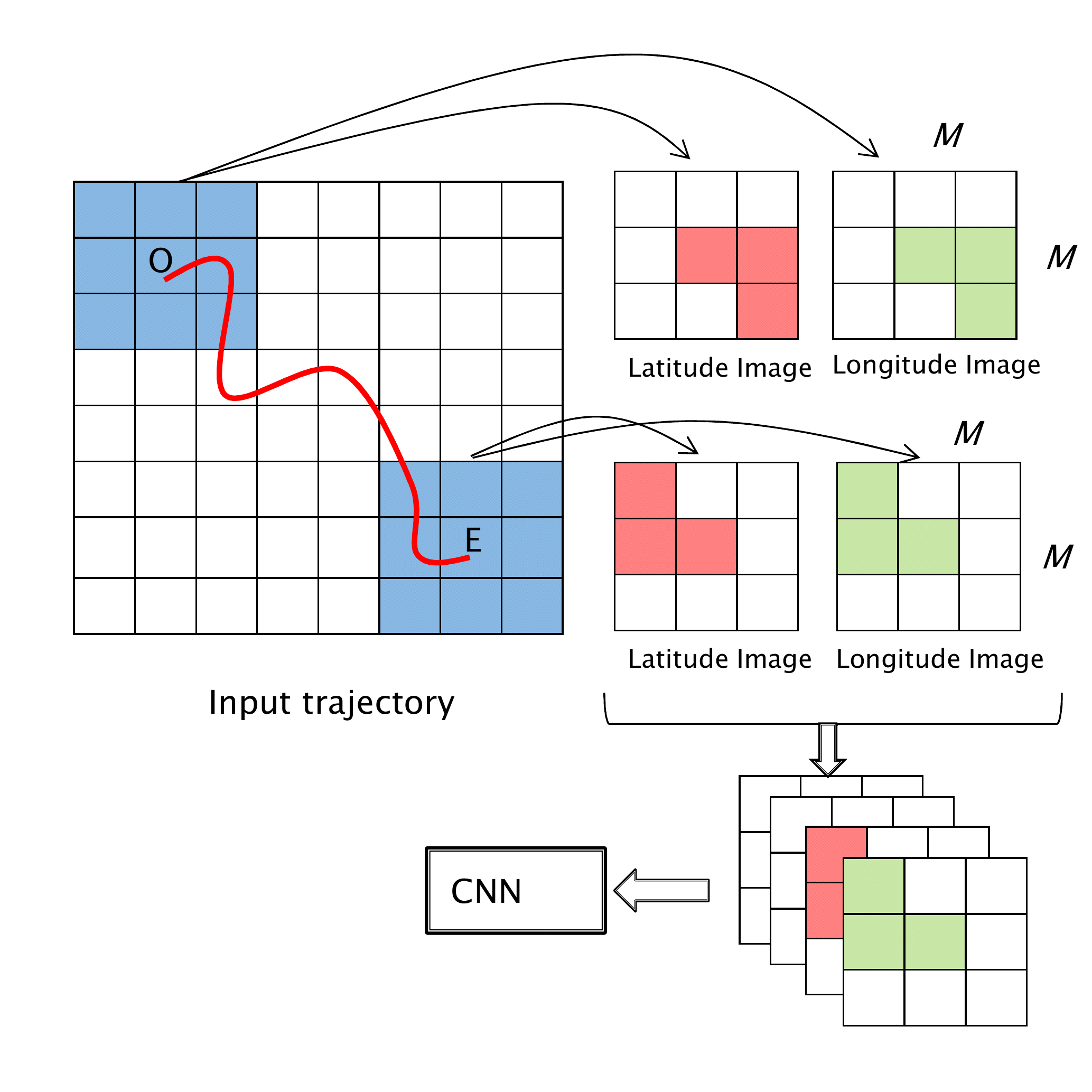}
\end{minipage}
}
\caption{The architecture of the local enhancement CNN. In this model, the two local areas which are close to the start location 'O' and the end location 'E' of the trajectory are convolved deeply.}
\label{T-CONV-local}
\end{figure}

Based on the observation that the local portions near the start and end of a trajectory play more significant roles to the destination prediction, we can convolve these local areas deeply beyond deploying convolution over the global image. Fig.~\ref{T-CONV-local} shows the architecture of this model, which focuses on the two local areas centered at the start and the end of an input trajectory.  Each local area is then divided into an $M*M$ grid and recorded as two $M*M$ trajectory images: \emph{latitude image} $I_a$  and \emph{longitude image} $I_o$, which record the precise latitudes and longitudes of the points in the sub-trajectory belonging to the local area. Specifically, given the local area, which contains the sub-trajectory $\zeta_{i,j}' = <\delta_{i,j,1}', \delta_{i,j,2}',...,\delta_{i,j,n}'>$, the pixel value of its latitude image $I_a$ is assigned as:

\begin{small}
\[I_{a}(m,n)=\begin{cases}
Norm(\Phi(C_{m,n}))&\text{if $\exists \delta_{i,j,k}'(\delta_{i,j,k}' \triangleright C_{m,n})$ }\\
0&\text{otherwise}
\end{cases}\]
\end{small}.
The pixel value of its longitude image $I_o$  is assigned as follows:
\begin{small}
\[I_{o}(m,n)=\begin{cases}
Norm(\Upsilon(C_{m,n}))&\text{if $\exists \delta_{i,j,k}'(\delta_{i,j,k}' \triangleright C_{m,n})$}\\
0&\text{otherwise}
\end{cases}\]
\end{small}.
Here $\Phi(C_{m,n})$ means the latitude of the center of the cell $C_{m,n}$, while $\Upsilon(C_{m,n})$ means the longitude of the same position. $Norm(.)$ denotes the normalization function to transform its input into the range [-1,1]. With the above processing, totally four images can be obtained, which are stacked together and fed into the convolutional neural network of Fig.~3.

The local enhancement convolution on the important local areas may benefit efficient extraction of important patterns for prediction. Furthermore, by only focusing on small subsets of the whole trajectory image, the overlapping degrees of input latitude images and longitude images among different trajectories are much higher than original global trajectory images. This can help overcome the sparsity problem of trajectories \cite{xue2015solving}, and facilitate the convergence of the learning procedure.

\section{Evaluation}
In this section, we conduct experiments to validate the effectiveness of the T-CONV models, and compare them with the other state-of-the-art algorithms.

\subsection{Experiment setup}

We test the performance of T-CONV on the real trajectory dataset of the ECML-PKDD competition\cite{kaggle2015}, which  is collected from 442 taxis running in the city of Porto for a complete year (from 2013-07-01 to 2014-06-30), and has totally 1.7 millions complete trajectories.  In order to validate the prediction accuracy, we use the same testing dataset as \cite{de2015artificial}, which  contains 19,770 incomplete trajectories randomly selected from the original training set.  The remaining trajectories are used for training.

Two T-CONV models are implemented: the \emph{T-CONV-Basic} which means the simple  CNN model presented in section IV.B, and the \emph{T-CONV-LE} which means the local enhancement CNN model mentioned in section IV.D. We mainly compare T-CONV with the neural networks based models\cite{de2015artificial}, which win the champion of the ECML-PKDD competition\cite{kaggle2015}. These models are listed as follows:

 \begin{table}[h]
 \centering
 \caption{The evaluation error of models} \label{tab:notation}

 \begin{tabular}{p{5.0cm}| p{1.5cm}}
 \toprule
 Model & Error \\
 \hline
 \hline
  MLP & 2.81 \\
\hline
Bidirectional RNN & 3.01 \\
\hline
    Bidirectional RNN with window & 2.60 \\
\hline
    Memory network  & 2.87 \\
\hline
\hline
    \textbf{T-CONV-Basic} & \textbf{2.70} \\
\hline
    \textbf{T-CONV-LE} & \textbf{2.53} \\
 \bottomrule
 \end{tabular}
 \end{table}

\begin{itemize}
\item \emph{MLP}: a simple multi-layer perceptron model. The input includes the latitudes and longitudes of the first and last 5 points of an input trajectory.

\item \emph{Bidirectional RNN}: a bidirectional recurrent neural network which considers a trajectory as a sequence, and  each GPS point in the trajectory forms a transition state of RNN.

\item \emph{Bidirectional RNN with window}: a variant of the above Bidirectional RNN model, which uses a sliding window of 5 successive GPS points as a transition state of RNN.

\item \emph{Memory network}: another variant of the above RNN model which encodes the trajectories into vectors with fixed length, and compares the similarity between the input trajectories and historical trajectories in this vector space.
\end{itemize}

The evaluation error to measure these models is the Haversine distance between the real destination and the predicted location. The distance is defined as:
\begin{small}
 \begin{eqnarray}
   d(x,y) &=& 2 \cdot r \cdot \arctan (\sqrt{\frac{a}{1-a}})
\end{eqnarray}
\end{small}
where
\begin{small}
 \begin{eqnarray}
   a = \sin^{2}(\frac{\phi_x - \phi_y}{2})+\cos£¨\phi_x£©\cos£¨\phi_y£©\sin^{2}(\frac{\lambda_x - \lambda_y}{2})
\end{eqnarray}
\end{small}
, $\phi$ is the latitude, $\lambda$ is the longitude and  $r$  is the radius of the earth.

\subsection{Performance}

The prediction errors of our proposed models are compared with the baseline models listed in Table 1. The result shows that both T-CONV models are better than the baseline MLP models. As shown in Fig.~\ref{cnn}, T-CONV models share similar output layers as the MLP layers, but use the multi-layer convolutional neural networks to capture  trajectory patterns. The superior performance of T-CONV demonstrates that the effectiveness of  this kind of multi-scale patterns extracted by T-CONV. In addition, the T-CONV models are also better than the Bidirectional RNN model and Memory network, which are also based on one spatial scale.

\begin{figure}
\centering
{
\begin{minipage}[b]{0.4\textwidth}
\includegraphics[width=1\textwidth]{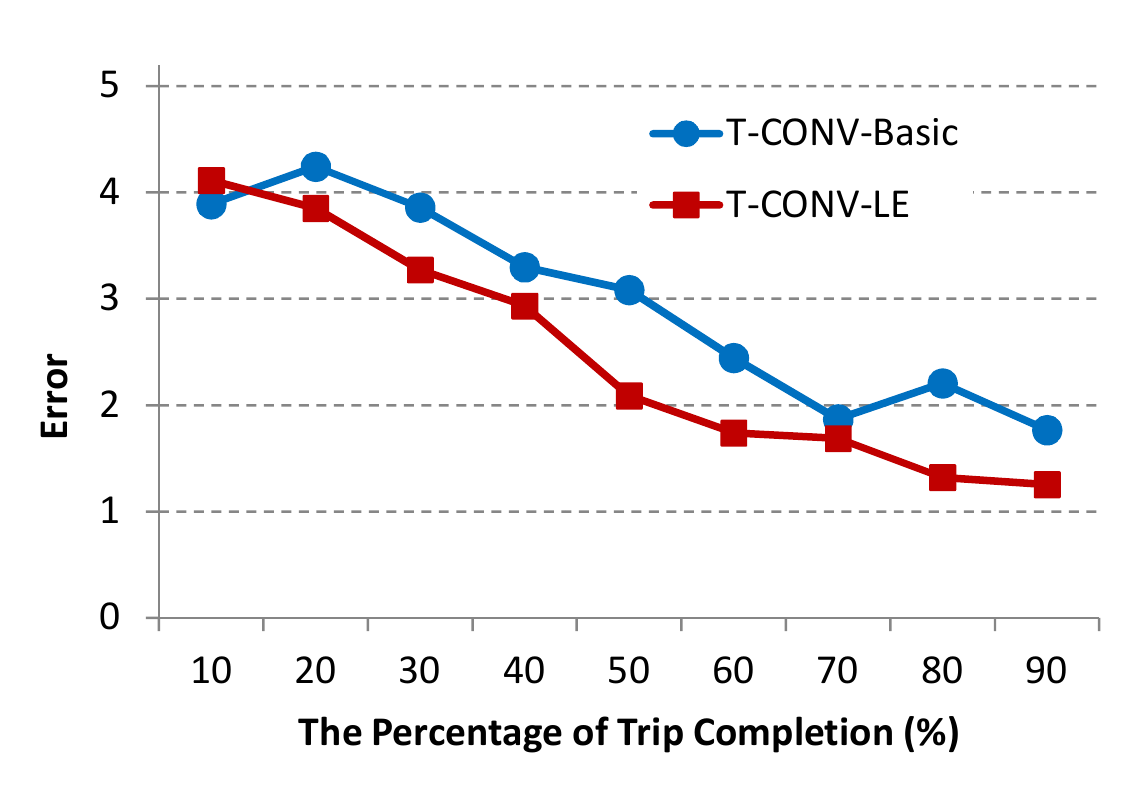}
\end{minipage}
}
\caption{The prediction error given different completeness ratio of trajectories }
\label{finish-ratio}
\end{figure}

When looking into the two T-CONV models, T-CONV-LE is much better than T-CONV-Basic, which  shows that the convolution on the important local areas can enhance the ability of CNN to capture significant patterns. As shown in Fig.~\ref{finish-ratio}, we also test the prediction performance of these two models given the trajectories with different completeness ratios. The completeness ratio indicates the percentage of the current observed trajectory compared to its  full trajectory, and it implies how far the end point of the trajectory to destination. Fig.~\ref{finish-ratio} shows that  the closer of the taxi to the destination, the lower prediction error to be achieved. Furthermore, it also shows that T-CONV-LE performs much better in most of the cases.

\subsection{Sensibility of parameters}

 In this section, we test  how the change of the hyper parameters in T-CONV-LE may affect its prediction efficiency.  We focus on the two parameters related to the spatial scale of trajectory processing, namely, M and W:

 \begin{itemize}
 \item $M$: As shown in Fig.~\ref{T-CONV-local}, $M$ is the number of rows (or columns) of the grid. A Higher $M$  corresponds to a denser grid with small granularity of the cells.
\item $W$: Measured in meters, $W$ is the width of each cell of the grid.  Thus $M*W$ indicates the size of the local areas of T-CONV-LE.
 \end{itemize}

\begin{figure}
\centering
{
\begin{minipage}[b]{0.4\textwidth}
\includegraphics[width=1\textwidth]{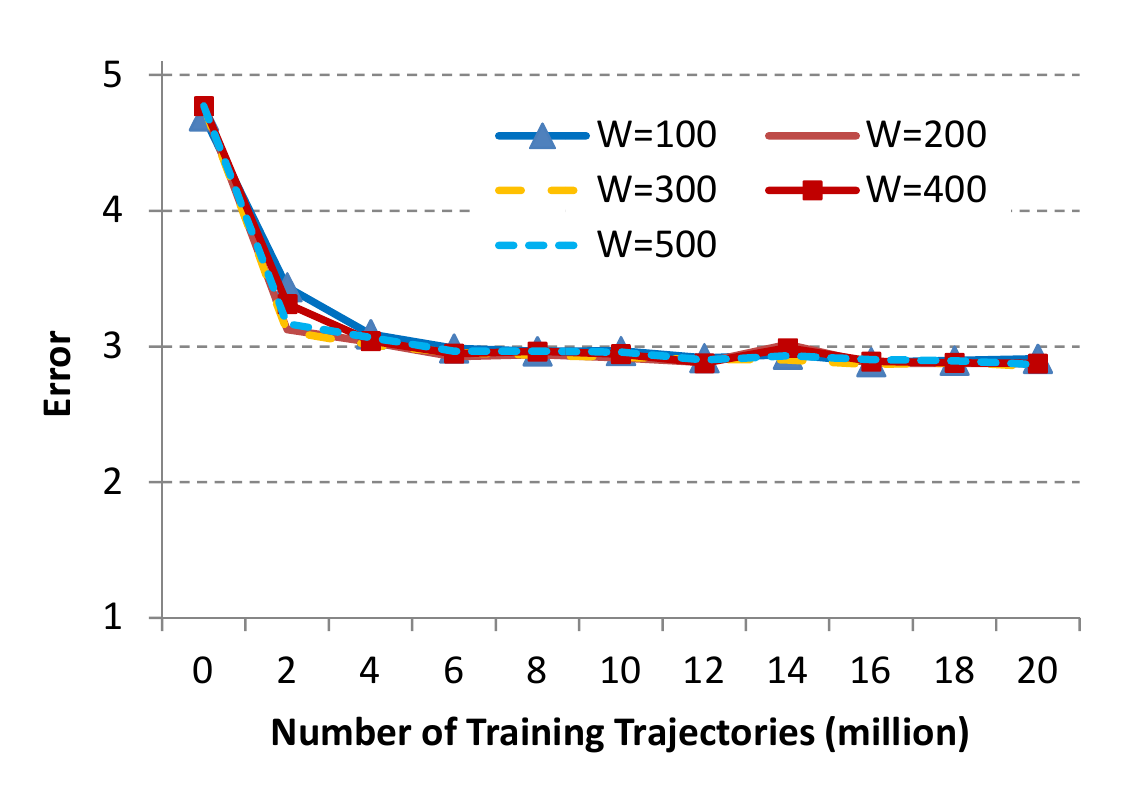}
\end{minipage}
}
\caption{The prediction error of the T-CONV-LE model with different $W$, which is the width of each cell. $M$ is kept fixed as 30 here.}
\label{W}
\end{figure}

\begin{figure}
\centering
{
\begin{minipage}[b]{0.4\textwidth}
\includegraphics[width=1\textwidth]{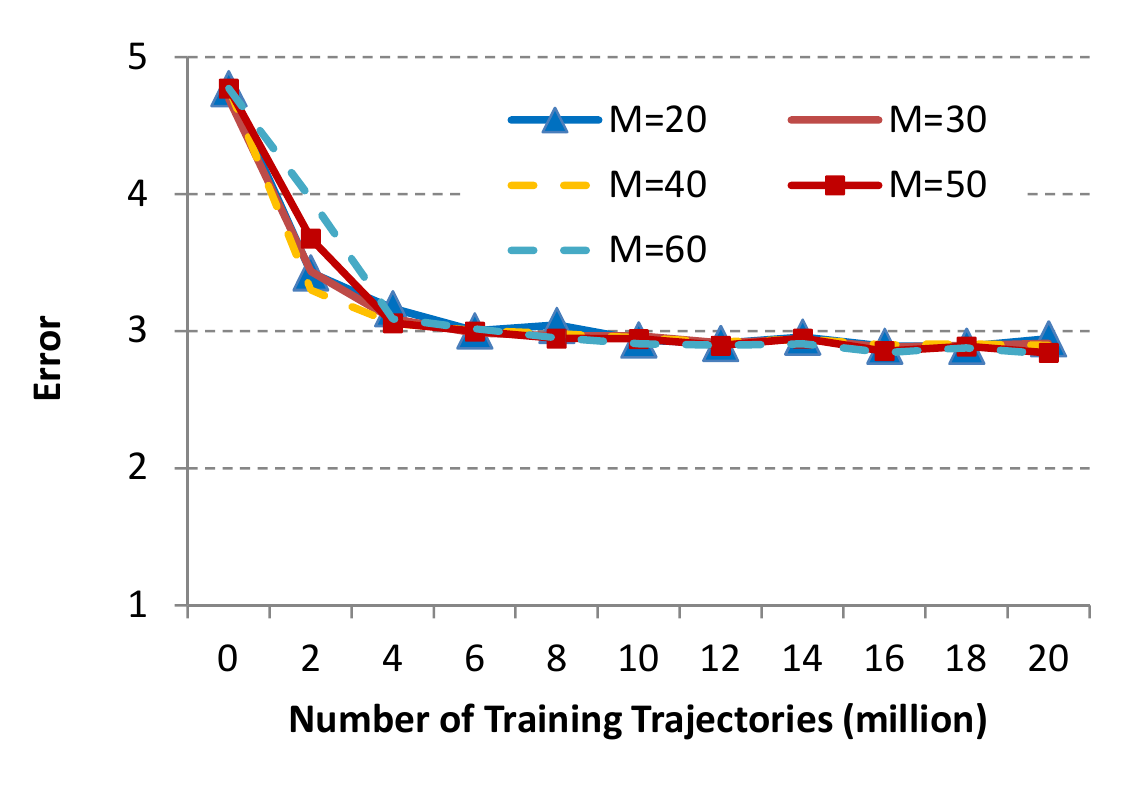}
\end{minipage}
}
\caption{The prediction error given different $M$, which is the number of the rows in the grid. Here $W$ is kept fixed as 100.}
\label{M}
\end{figure}

We test the performance of  T-CONV-LE with different combinations of various $W$ and $M$. Fig.~\ref{W} and Fig.~\ref{M} show how the prediction error decreases during the training stage. Specifically, Fig.~\ref{W} shows the performance of  the models with different $W$ while $M$ is unchanged, and Fig.~\ref{M} shows the performance of  the models  with different $M$ given fixed $W$. Both Fig.~\ref{W} and Fig.~\ref{M} indicate that the performance of T-CONV-LE is not sensitive to the slight change of $W$ and $M$. Thus, the multi-layer convolutional neural networks of the model are robust to extracting multi-scale patterns for precise prediction.

\section{ Conclusions}
In this paper, we have presented a convolutional neural network based model, T-CONV, to predict the destinations of taxi trajectories. Different from traditional research works which process trajectories as one-dimensional sequences in one single scale, T-CONV models trajectories as two-dimensional images and combines multi-scale trajectory patterns through multi-layer convolution operations. Furthermore, inspired by the observation that different portions of a trajectory have significant difference of contribution to the final prediction, a local enhancement T-CONV model named T-CONV-LE is presented to efficiently extract the patterns of the important local areas in a trajectory. Experiments show that T-CONV-LE can achieve better prediction accuracy than state-of-the-art algorithms.

In the future, we will extend the current T-CONV-LE model, which has two local enhancement areas with fixed size, to the more general one, which has multiple local enhancement areas with tunable size. We will also extend the experiments to validate our model in more real trajectory datasets.

\bibliographystyle{unsrt}

\end{document}